\documentclass{article}

\usepackage{PRIMEarxiv}

\usepackage[utf8]{inputenc} 
\usepackage[T1]{fontenc}    
\usepackage{hyperref}       
\usepackage{url}            
\usepackage{booktabs}       
\usepackage{amsfonts}       
\usepackage{nicefrac}       
\usepackage{microtype}      
\usepackage{lipsum}
\usepackage{fancyhdr}       
\usepackage{subcaption}
\usepackage{graphicx}       
\graphicspath{{media/}}     

\title{Artificial Perceptual Learning:\\ Image Categorization with Weak Supervision}

\author{%
  Chengliang Tang \\
  Department of Statistics\\
  Columbia University\\
  New York, NY 10027 \\
  \texttt{ct2747@columbia.edu} \\
   \And
   Mar\'ia Uriarte \\
   Department of Ecology, Evolution \\
   and Environmental Biology \\
   Columbia University \\
   New York, NY 10027 \\
   \texttt{mu2126@columbia.edu} \\
   \And
   Helen Jin \\
   Department of Computer \\
   and Information Science \\
   University of Pennsylvania \\
   Philadelphia, PA 19104 \\
   \texttt{helenjin@seas.upenn.edu} \\
   \And
   Douglas C. Morton \\
   Goddard Space Flight Center \\
   NASA \\
   Greenbelt , MD 20771 \\
   \texttt{douglas.morton@nasa.gov} \\
   \And
   Tian Zheng \\
   Department of Statistics\\
   Columbia University\\
   New York, NY 10027 \\
   \texttt{tz33@columbia.edu} \\
}

\begin{document}
\maketitle

\begin{abstract}
Machine learning has achieved much success on supervised learning tasks with large sets of well-annotated training samples. However, in many practical situations, such strong and high-quality supervision provided by training data is unavailable due to the expensive and labor-intensive labeling process. Automatically identifying and recognizing object categories in a large volume of unlabeled images with weak supervision remains an important, yet unsolved challenge in computer vision. In this paper, we propose a novel machine learning framework, {\em artificial perceptual learning} (APL), to tackle the problem of weakly supervised image categorization. The proposed APL framework is constructed using state-of-the-art machine learning algorithms as building blocks to mimic the cognitive development process known as {\em infant categorization}. We develop and illustrate the proposed framework by implementing a wide-field fine-grain ecological survey of tree species over an 8,000-hectare area of the El Yunque rainforest in Puerto Rico. It is based on unlabeled high-resolution aerial images of the tree canopy. Misplaced ground-based labels were available for less than 1\% of these images, which serve as the only weak supervision for this learning framework. We validate the proposed framework using a small set of images with high quality human annotations and show that the proposed framework attains human-level cognitive economy. \footnote{An applied version of this paper is published as "Tang, C., Uriarte, M., Jin, H., C Morton, D., \& Zheng, T. (2021). Large‐scale, image‐based tree species mapping in a tropical forest using artificial perceptual learning. \emph{Methods in Ecology and Evolution}, 12(4), \href{https://besjournals.onlinelibrary.wiley.com/doi/10.1111/2041-210X.13549}{608-618}."}
\end{abstract}

\section{Introduction}
Over the past decade, machine learning algorithms, especially those with deep structures, have led to breakthroughs in supervised learning. Tasks such as image classification \cite{he2016deep, krizhevsky2012imagenet} and object segmentation \cite{long2015fully} in computer vision have seen substantial improvements in performance. In order to attain such performance, the training of most state-of-the-art supervised learning algorithms requires a large volume of well-annotated data. For example, the popular benchmark dataset PASCAL VOC \cite{everingham2010pascal} provides accurate image labels, bounding boxes and pixel-wise segmentation information for a variety of object categories. Under the {\em strong supervision} provided by high-quality training data, end-to-end deep neural networks can be readily built and evaluated for computer vision tasks \cite{liu2016ssd, ren2015faster}.

\begin{figure}
\begin{subfigure}{0.48\textwidth}
\begin{center}
\includegraphics[width=0.9\linewidth]{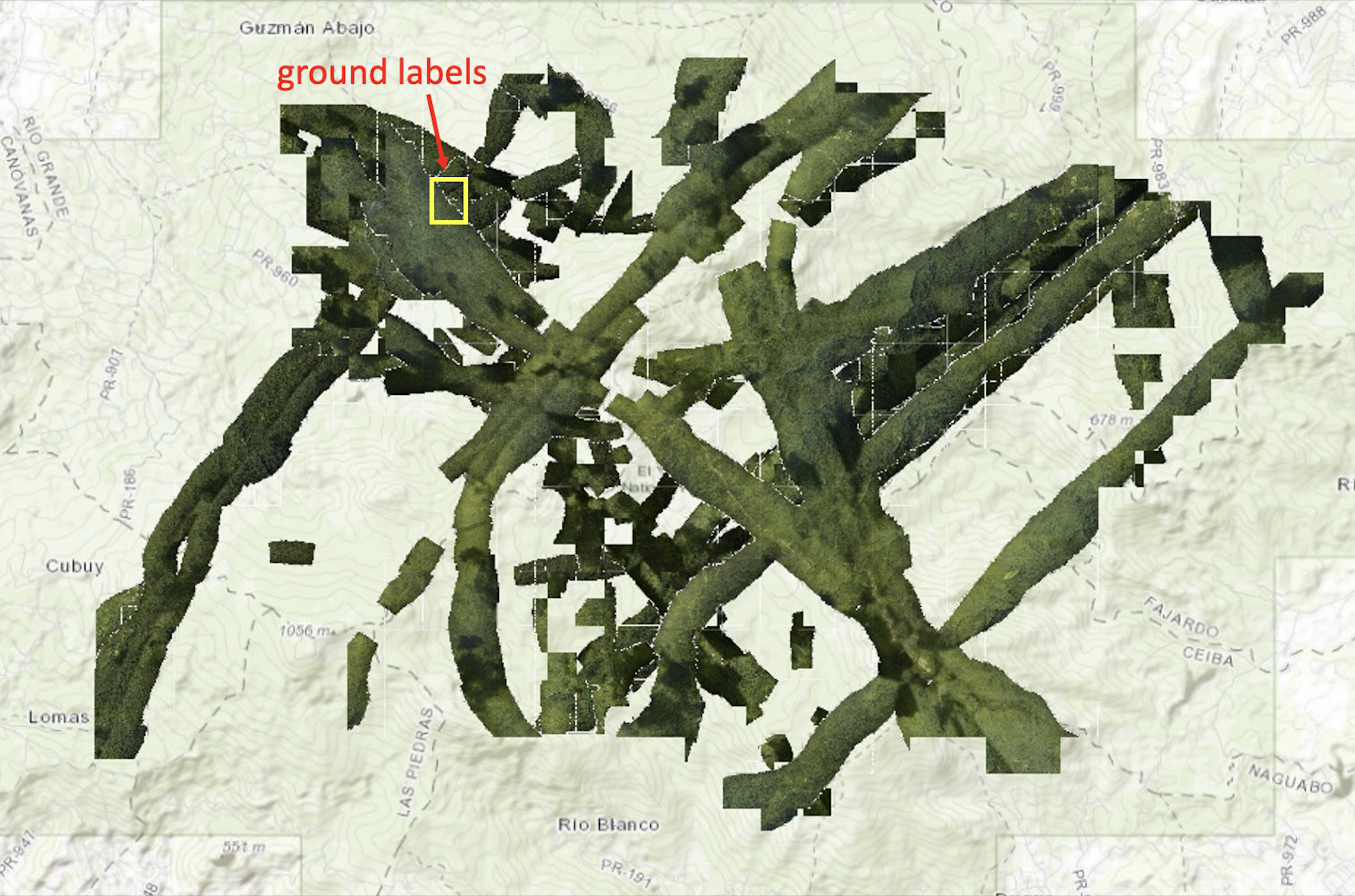}
\end{center}
\caption{Map of the El Yunque rainforest in Puerto Rico showing aerial images collected. Tree species labels at trunk-locations by ground-based observation are available in the yellow bounding box.} \label{fig:ground-1a}
\end{subfigure}
\hspace*{\fill} 
\begin{subfigure}{0.48\textwidth}
\begin{center}
\includegraphics[width=0.9\linewidth]{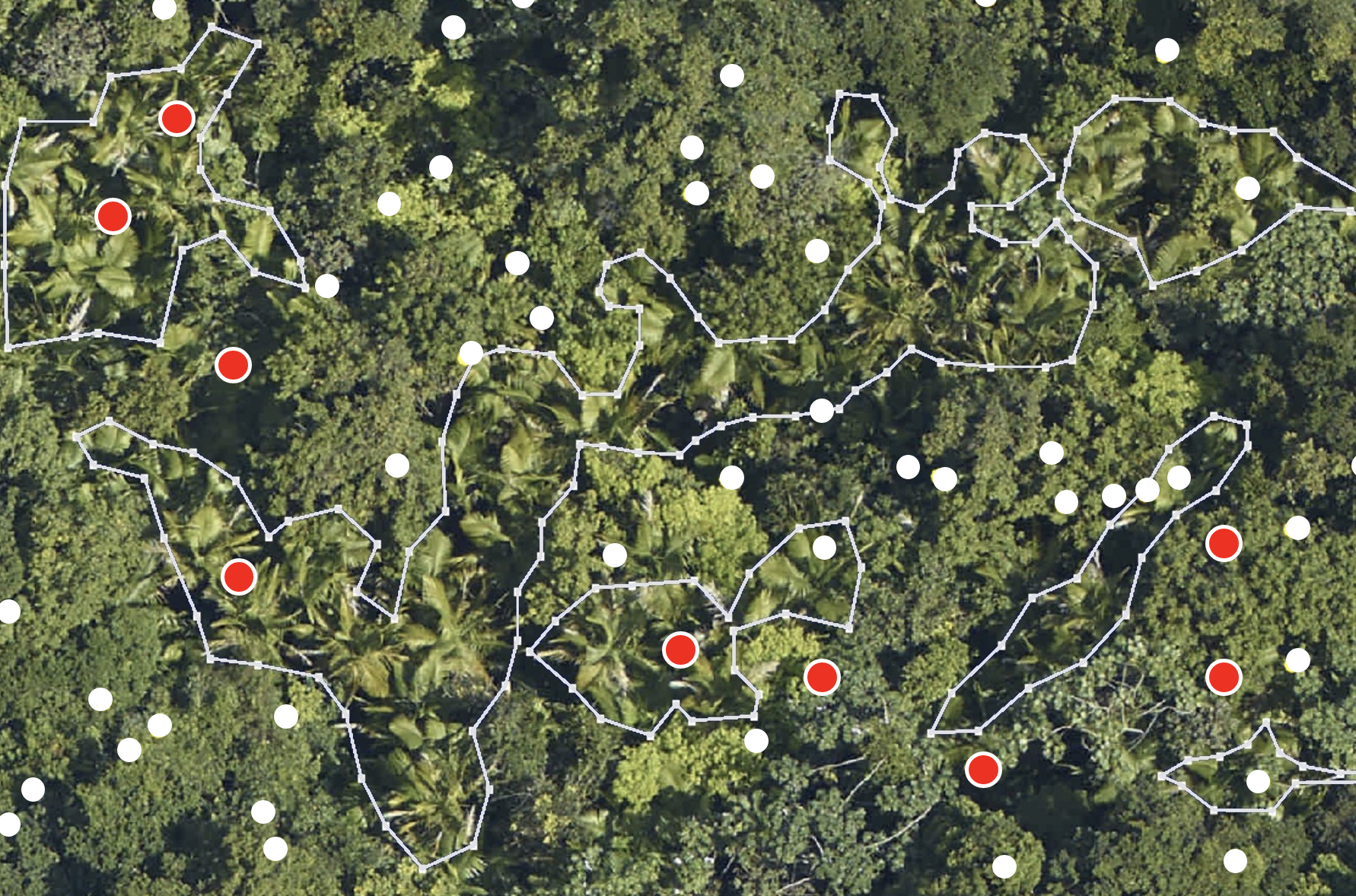}
\end{center}
\caption{A high resolution image of tree canopies with mismatched ground labels (large red dots for the palm trees and small white dots for others). Palm tree canopies are segmented using closed polygonal lines.} \label{fig:ground-1b}
\end{subfigure}
\caption{Unlabeled/mislabeled aerial images of tree canopies call for machine learning algorithms with weak supervision. } \label{fig:dataset}
\end{figure}

Nonetheless, in most real-world applications, such strong supervision is unavailable due to the expensive and labor-intensive labeling process. Recently developed technologies such as fMRI and remote sensing have enabled efficient gathering of high-quality unstructured image data. However, human expert knowledge that is critical for accurate data annotation remains expensive. As a result, labeled training sets in these situations are often partially available and rife with inaccuracies, providing, at best, only {\em weak supervision} to the learning tasks. Figure~\ref{fig:dataset} illustrates an example of weak supervision in a set of remote sensing data. High-resolution aerial images were captured in a large scale data collection over the El Yunque rainforest of Puerto Rico (Figure~\ref{fig:ground-1a}; see Section~\ref{sec_forest} for detail on this data set.). In the small rectangular area marked in Figure~\ref{fig:ground-1a}, tree species were annotated at the root/trunk locations of the trees by human workers, based on ground-based observation, which are referred to as {\em ground labels} throughout the rest of the paper. Trunk locations are known to have marked differences from the spatial distribution of the canopies of the same trees.  Figure~\ref{fig:ground-1b} displays the significant mismatch between ground labels and aerial images of the canopies. In this type of ecological study, there is a great need for machine learning algorithms that can automatically identify and recognize tree species based on crown characteristics with weak supervision.

In contrast to machines, humans are capable of abstracting general knowledge from multiple small-scale and noisy data sources \cite{carroll1993human, lake2017building}. When seeing a new animal species for the first time, human learners do not need to memorize numerous features by reviewing thousands of training samples. Instead, a simple conceptual prototype of that species would be characterized through an automatic, and often unconscious, cognitive process.
Such cognitive ability has been observed in human infants, even though their brains are regarded as underdeveloped and of limited memory \cite{bergelson20126, national2000people}. In particular, {\em categorization} is a primitive and important ability in all behavior and mental functioning \cite{quinn2002early}. Studies of visual preference and object examination reveal that infant categorization develops gradually through the process of quantitative enrichment \cite{gelman2011child, mareschal2001categorization, quinn1997emergence}. For a newborn infant, categories are initially formed at the basic level; superordinate categories develop as the infant begins to group together similar basic-level representations. In this process, parents/teachers' input (labeling and/or explanation) serves as a weak supervision.

In this paper, we propose {\em Artificial Perceptual Learning} (APL), a machine learning framework for image categorization with weak supervision. This framework is modeled after the cognitive {\em categorization} process of infants, with state-of-the-art machine learning algorithms as building blocks (Figure~\ref{fig:workflow}).
The rest of the paper is organized as follows. We first review related work on weakly supervised learning in Section 2 and introduce our APL framework in Section 3. In Section 4, we illustrate APL with an application to extracting information on tree species from unlabeled aerial images of tree canopies and provide numerical experiment results. We conclude with Section 5. 

\section{Related Work}
Recently, algorithms for weakly supervised learning have received considerable interest due to their applicability to many practical challenges. Following the framework of \cite{zhou2017brief}, tasks with weak supervision can be categorized into three main types: incomplete supervision, inexact supervision and inaccurate supervision. 

{\em Incomplete supervision} refers to the situation where labels are only available for a small subset of training data. To solve the problem of data insufficiency, active learning algorithms \cite{settles2009active} attempt to better extract label information by ``actively'' asking an "oracle" (e.g., a human annotator) for queries of selected unlabeled instances. This framework has been widely used in image classification \cite{joshi2009multi, kapoor2007active, li2013adaptive}. For many applications, however, a human expert (oracle) might be too expensive for the time-intensive training process. Meanwhile, semi-supervised learning algorithms \cite{chapelle2009semi, zhu2005semi} utilize the unlabeled training data as well as labeled data to improve prediction accuracy. In both cases, noise-free labeled data are of great importance for model performance.

{\em Inexact supervision} addresses the situation where the given labels are at coarser scales than desired. For example, in many real-world object segmentation tasks, only image-level training labels are available, while the task is to localize each object. Multi-instance learning \cite{zhou2007multi} was proposed to address this challenge with the {\em bag-of-instances} setup: instances are organized in bags, of which the labels are given in the training set. In document classification, concept labeling \cite{chenthamarakshan2011concept} built a Bayesian framework connecting the (finer) document labels with the (coarser) concept labels, which also significantly decreased labeling cost.

{\em Inaccurate supervision} concerns the situation where labels are a noisy version of the ground truth. To learn with noisy labels, many algorithms make the assumption that the noises are randomly generated. \cite{brodley1999identifying} proposes to first identify the potentially mislabeled instances and perform label correction. The data programming approach proposed by \cite{ratner2016data} is a paradigm for integrating noisy labels from multiple sources, and creating a better training set using a dependency graph that can incorporate different data generating assumptions. 
\cite{inoue2017multi} proposes to manually label a small subset of training set, and then learn the nonlinear mapping from noisy labels to ground truth using a convolution neural net. 

{\em Transfer learning} \cite{pan2010survey, weiss2016survey}, another approach for weakly supervised learning, has seen a lot of recent development when applied to the aforementioned three weak supervision situations \cite{gao2014transfer, raina2007self}. Relaxing the usual assumption of independent and identically distributed training data, transfer learning assumes a common information processing pipeline that can be shared between similar learning tasks, such as natural language processing and computer vision. Deep neural networks are particularly well-suited for transfer learning \cite{bengio2011deep}. The relationship between transferability and network structure was empirically explored in \cite{mou2016transferable, ozbulak2016transferable, yosinski2014transferable}.

For many real-world learning tasks, all the above weak supervision scenarios may apply at the same time (e.g., noisy and inexact labels are only available on a small subset, as seen in Figure~\ref{fig:dataset}).
In addition, the required setups (e.g., a human oracle who is on call, or high-quality labels on bags of instances) for existing weakly supervised learning frameworks are usually not met due to the complexity of real-world applications. A scalable and flexible learning framework that can be generally applied to different tasks is highly desirable.

\section{Artificial Perceptual Learning (APL)}
\begin{figure}[h]
	\includegraphics[width=0.9\textwidth]{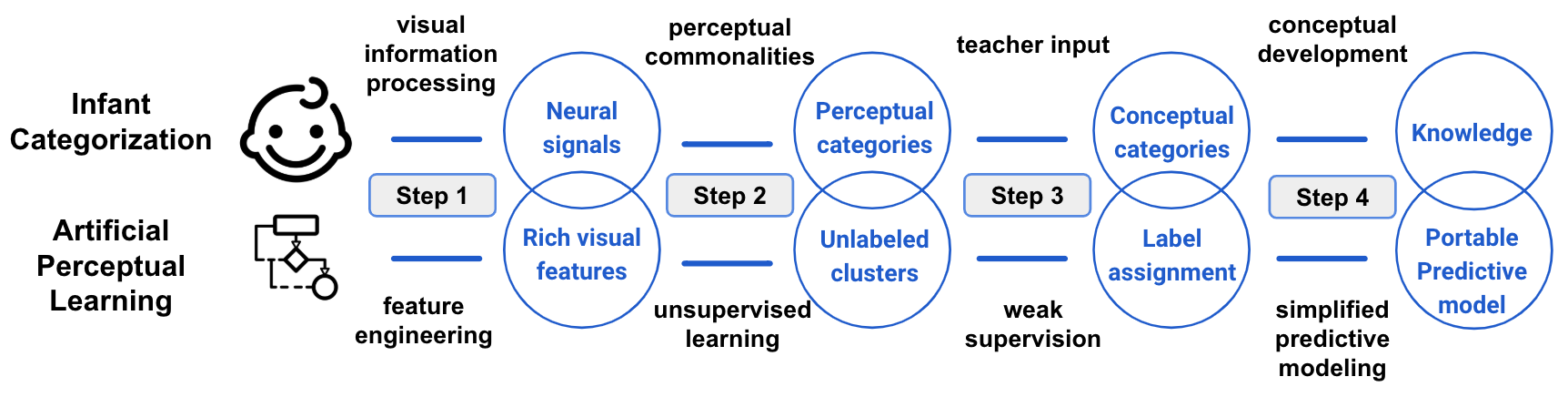}
	\centering
	\caption{Artificial perceptual learning (APL) versus the cognitive process of infant categorization.}
	\label{fig:workflow}
\end{figure}

In computer vision, the form of weak supervision might be complicated and specific to a particular task. Thus, it is impractical to address the challenge of weak supervision with a single machine learning algorithm. We propose \emph{artificial perceptual learning} (APL), a {\em workflow} framework for learning image categorization with weak supervision, which models the cognitive process of human learners to achieve better learning efficiency (Figure \ref{fig:workflow}). 


In the proposed framework, state-of-the-art machine learning algorithms could be adopted as building blocks to achieve better sample efficiency. The main idea is to create a large high-quality training set using the first three steps in Figure~\ref{fig:workflow}, and then build stable and portable predictive models through conventional supervised learning. 

{\bf Step 1: visual feature engineering. } The first step in the proposed framework is to derive what a computer ``see'' in an image. 
Here, the goal of visual feature engineering is two-fold: to remove random and non-random noises, such as denoising and shadow removal, which is application specific; and to process unstructured arrays of pixels of an image into structured vectors of rich and meaningful visual features.
%
This visual feature engineering step in APL is similar to
the bottom-up processing structure of human's visual system
\cite{gibson1966senses}. Visual stimuli received by our eyes are processed into neural signals by the visual nervous system (VNS), and interpreted by our brains. The visual attention theory \cite{bundesen1990theory} from cognitive psychology indicates that much of irrelevant information is filtered out during this process. The visual nervous system's ability to carry out this information filtering is partly due to the biological structure of the visual system, as a result of millions of years of human evolution. As an infant grows and learns, the visual nervous system is further trained based on experience. Therefore, when humans see an object for the first time, our visual nervous system processes the visual inputs based on how our brain has been attuned to categorize objects that we have seen. 


Existing general-purpose {\em pretrained deep networks} in computer vision, such as VGG network \cite{simonyan2014very}, ResNet \cite{he2016deep}, and AlexNet \cite{krizhevsky2012imagenet}, are well suited for learning expressive image representations in this step, as they have achieved excellent generalization performance in transfer learning \cite{ng2015deep, shin2016deep}. In particular, they have been shown to offer the utility of mapping different visual patterns, e.g., color, shape, textures, into linearly separable classes in the feature space \cite{guerin2017cnn, xie2016unsupervised}.

{\bf Step 2: unsupervised learning.} Based on the extracted visual features,
the goal of the Steps 2\&3 is to derive labeled prototypes with weak supervision. In Step 2, image samples are organized into disjoint groups, or {\em prototypes}, based on their visual similarity using unsupervised learning algorithms. In Step 3, these prototypes would be evaluated and assigned {\em bag}-level labels using information from weak supervision.
%
These two steps are the key components in the APL workflow, an idea that originates from our understanding of infant categorization \cite{gelman2011child, mareschal2001categorization, quinn1997emergence}. Studies \cite{mareschal2001categorization, murphy2004big} have shown that even infants under one year were able to spontaneously recognize ``basic-level'' categories, known as {\em perceptual categories}, with little supervision. These perceptual categories are simple groupings of visually similar stimuli. An infant assembles these grouping according to perceptual commonalities (e.g., shape, color, and etc.) without understanding their significance. These primitive perceptual categories provide the basis for a cognitive structure to incorporate supervision and form advanced and abstract categories.

In the proposed APL framework, the cognitive process of forming perceptual categories is replaced by applying unsupervised learning algorithms on the extracted image features. Deep features from  pretrained models are expected to map visually similar image samples into proximate neighborhoods in the feature space. Conventional clustering algorithms, such as {\em K-means clustering}, are expected to efficiently and reasonably organize the unlabeled sample set into (disjoint) groups of images that have closely similar visual patterns. Soft clustering methods, such as Gaussian mixture models (GMM), and hierarchical clustering that introduces more structured grouping, can also be considered, depending on the learning task at hand. 

{\bf Step 3: label assignment.}
Assigning labels to image categories/groups formed in Step 2 is the part of APL that requires weak supervision. In Figure \ref{fig:workflow}, this step corresponds to a parent/teacher's input in the process of infant categorization by providing imprecise information on a collection of mixed perceptual categories. 
With this weak supervision, infants are able to turn perceptual categories into knowledge of conceptual prototypes. In the APL framework, weak supervision may come in different forms
depending on the setup of the task. For example, in the case of inaccurate supervision (noisy labels), a practical solution is to calculate an ``average'' label for an identified cluster. If the information on the true labels is reflected in the features (Step 1), and leads to meaningful clusters (Step 2), this ``average'' label assignment is expected to mitigate noises from inaccurate supervision. 
This idea of prototype identification is similar to that of multi-instance learning \cite{zhou2007multi}, where the constructed clusters from Step 2 can be viewed as ``bags'' of instances. 
%



{\bf Step 4: predictive modeling.} Labeled clusters from Step 3 are now ready to be used as a training set for either conventional predictive modeling algorithms or the multi-instance learning framework.
%
This step corresponds to the post-infant conceptual development. When learning to categorize new objects, a more developed brain does not merely rely on simple perceptual categories based on basic visual characteristics. Instead, it will take advantage of previously learned concepts and category structures as meta features in a more supervised setting.

\section{Experiments}
To evaluate the APL framework's ability of learning with weak supervision, we analyze an extensive airborne collection of fine-resolution imagery that includes an ecological survey of tree species over the El Yunque rainforest in Puerto Rico. The computer vision task is to create tree canopy segmentations using unlabeled high-resolution aerial images. Imprecise ground locations of tree trunks serve as the only weak supervision. 
%
The machine learning visual perception workflow developed for this task is composed of four main steps following the APL framework: data preprocessing, visual feature engineering, prototype identification, and pixel-wise classification. 
We apply the proposed APL framework to obtain a map of palm tree distributions, and validate the result using a small set of images labeled by Amazon MTurk human annotators. 
\label{sec_forest}

{\bf Datasets.} In March 2017, high-resolution (3cm $\times$ 3cm) aerial images were captured at the landscape scale (area $> 8,000$ hectares) over El Yunque rainforest in Puerto Rico by \textit{NASA Goddard's LiDAR, Hyperspectral, and Thermal (G-LiHT) Airborne Imager}\footnote{https://glihtdata.gsfc.nasa.gov/puertorico/index.html}. In total, this dataset is composed of $962$ unlabeled full-color TIFF images, each of size $10,000$ \texttt{px} $\times$ $10,000$ \texttt{px} (spatial coverage = 300m $\times$ 300m). Ground observation data were collected in 2016 for a 16-hectare area of the G-LiHT coverage of the forest as part of the \textit{Luquillo Forest Dynamics Plot (LFDP).} They provide detailed information about tree characteristics, such as trunk locations, tree species, and trunk diameters. For experiments in this paper, we assembled small ($<1\%$) training sets that consists of images with LFDP ground labels, and potentially additional unlabelled images. 

{\bf Challenges.} In this experiment, the primary goal is to obtain a map of palm tree distributions using G-LiHT imagery and ground labels from the LFDP. The tropical setting of this scientific task creates a couple of computer vision challenges. The first challenge is the lack of a high contrast background. In tropical rainforests, trees grow in a dense neighborhood of other trees with similar canopy patterns. At the same time, overlapping canopies of large-leaf and small-leaf species would demand extra effort to obtain a fine-grain segmentation. The second challenge is the lack of accurately segmented training set. In most image segmentation tasks, a large training set with pixel-wise labels is provided for building end-to-end deep neural networks. In contrast, the ground labels in our task only exist in a small subset (less than $1\%$ of the images), and are in the form of scattered and imprecise ground locations of tree trunks that do not fully correspond to the imaging data of the tree canopies. 
In tropical rainforests, tree trunks often bend at sharp angles to gain greater sunlight exposure. As a result, significant misalignment can occur between unlabeled images and ground observations.

\begin{figure}[h]
	\begin{subfigure}{0.48\textwidth}
	\begin{center}
		\includegraphics[width=0.9\linewidth]{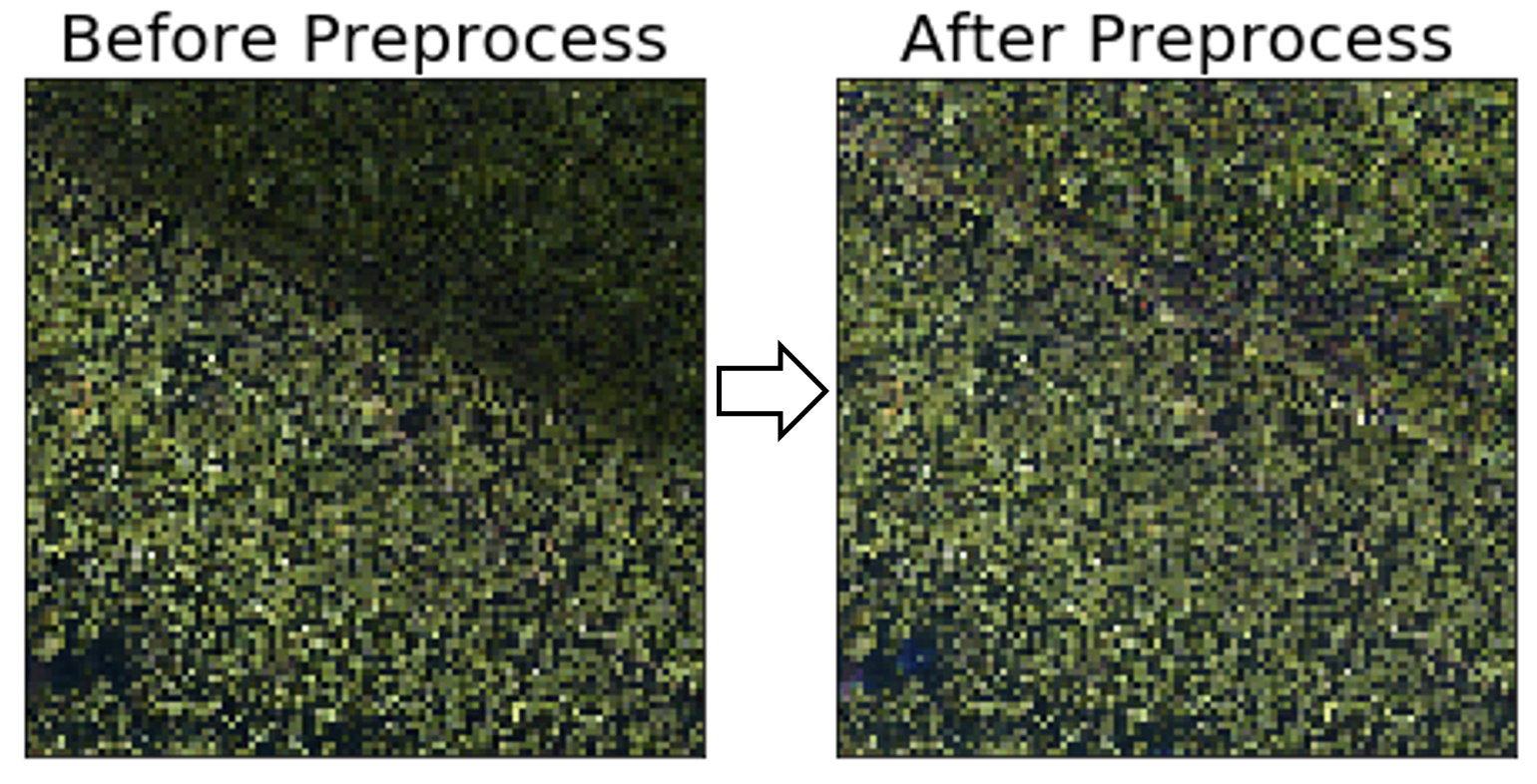}
	\end{center}
		\caption{One sample image} \label{fig:shadow-1a}
	\end{subfigure}
	\hspace*{\fill} 
	\begin{subfigure}{0.48\textwidth}
	\begin{center}
		\includegraphics[width=0.9\linewidth]{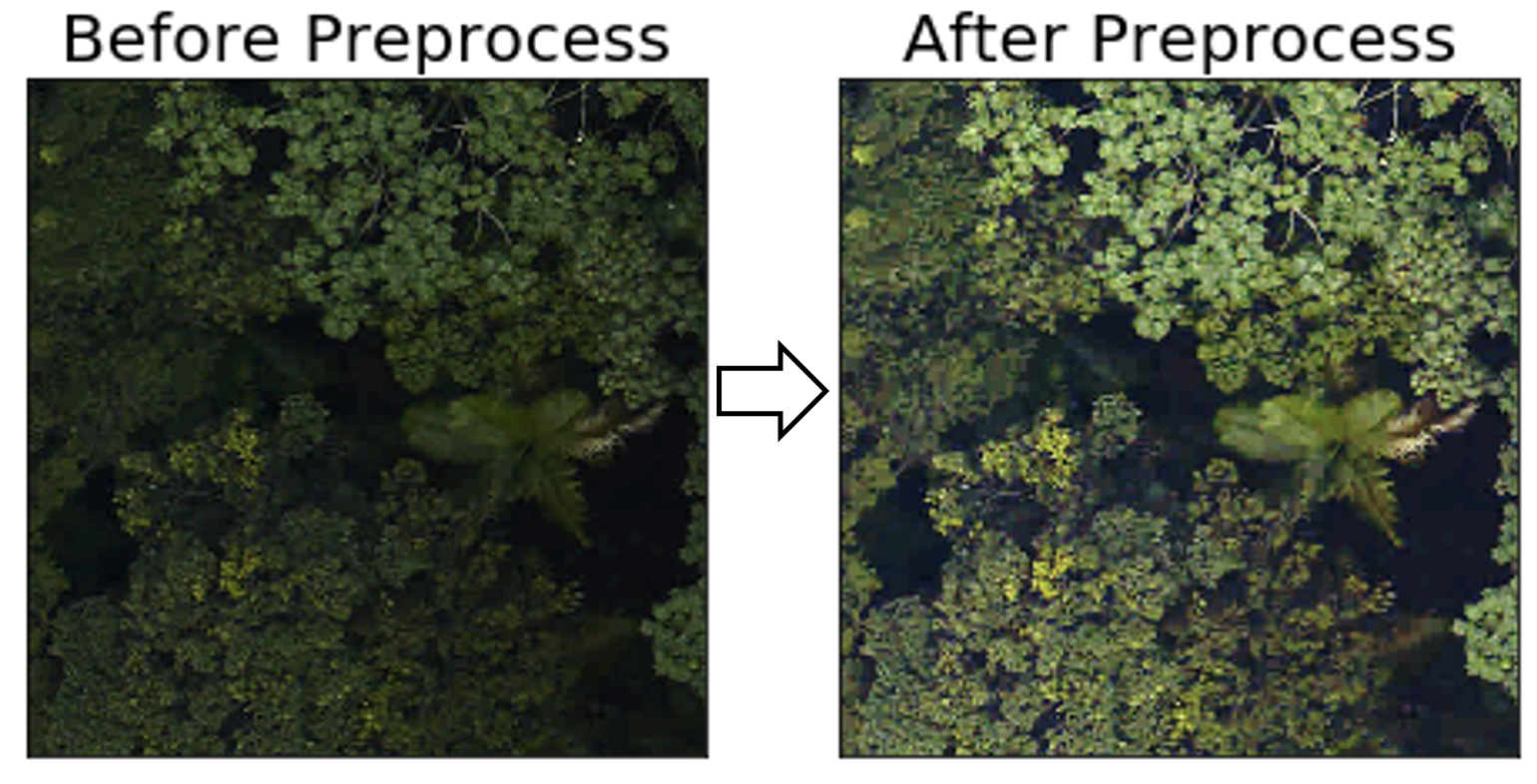}
	\end{center}
		\caption{Zoomed-in details} \label{fig:shadow-1b}
	\end{subfigure}
	\caption{Shadow detection and removal} 
	\label{fig:shadow-removal}
\end{figure}

\textbf{Data preprocessing.} Cloud shadows are common features in remote sensing data collected from tropical rainforests. The camera often cannot properly encode the scenes with spatially varying contrast. Uneven lighting due to shadows in images may lead to unstable and biased feature calculation. As sunlit areas account for a main portion of our dataset, using images without correcting for shadows will cause instability and systematic biases for the areas that were imaged in shadows. To achieve better prediction robustness, as part of data preprocessing, we implemented a shadow detection and removal procedure using common image analysis techniques \cite{bradski2008learning}. In the step of detection, we apply a low-pass filter to construct a smooth version of the grayscale image, followed by applying simple thresholding to create a mask of candidate shadow pixels. Then, in the step of shadow removal, histogram equalization is used in the RGB color space to adjust the contrast in shadow mask with reference to sunlit areas (Figure \ref{fig:shadow-removal}). In this part, hyperparameters are chosen by heuristic visual inspection.

\textbf{Visual feature engineering}. In this experiment, we used pretrained VGG-19 deep network as the feature extractor for the aerial images. Rather than directly feeding the whole $10,000$ \texttt{px} $\times$ $10,000$ \texttt{px} images into the deep network, we first cut the shadow-corrected images into small $100$ \texttt{px} $\times$ $100$ \texttt{px} patches, and fed them into the VGG-19 network. Outputs of the fully-connected layer were used as visual features. Here, the image patch size was chosen as a trade-off between prediction resolution and model variance. Smaller patch size can result in a better prediction resolution, while model variance would inevitably increase due to less informative visual patterns.

\textbf{Prototype identification with weak supervision.} Following the proposed APL framework, binary prototypes of {\em palm} versus {\em non-palm} were identified over image patches in two steps. First, a K-means clustering (k=$20$) model was trained over the VGG-19 features to generate unlabeled clusters. The goal of this step is to organize all the image patches into multiple disjoint groups according to their distances in the feature space, such that each group is consisted of semantically similar patches. 

\begin{figure}[h]
	\begin{subfigure}{0.48\textwidth}
		\includegraphics[width=\linewidth]{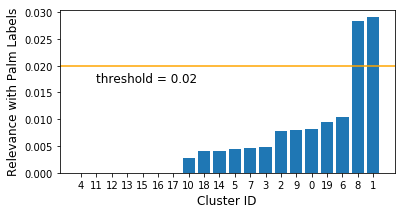}
		\caption{Palm relevance of each cluster} \label{fig:cluster-a}
	\end{subfigure}
	\hspace*{\fill} 
	\begin{subfigure}{0.48\textwidth}
		\includegraphics[width=\linewidth]{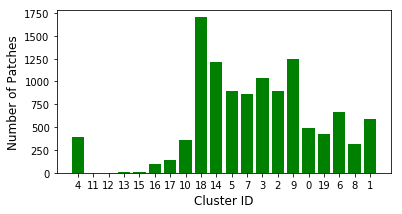}
		\caption{Size of each cluster} \label{fig:cluster-b}
	\end{subfigure}
	\caption{Palm relevance and size of each cluster} 
	\label{fig:cluster-density}
\end{figure}

With the unlabeled clusters in place, their relevance with palm trees were measured using training images with LFDP ground labels as shown in Figure \ref{fig:cluster-density}. As the image patches had been divided into disjoint groups, they provided a segmentation for the images with the LFDP ground labels. Tree species labels can be mapped to the identified clusters according to their ground trunk locations. 
In Figure \ref{fig:cluster-a}, palm relevance is defined as the number of palm observations in each cluster divided by the cluster size which is visualized in Figure \ref{fig:cluster-b}.

\begin{figure}[h]
    \centering
	\begin{subfigure}{0.22\textwidth}
	\begin{center}
		\includegraphics[width=0.9\linewidth]{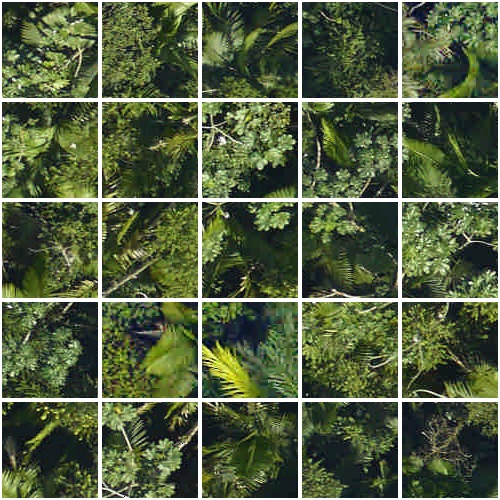}
	\end{center}
		\caption{Cluster 1} \label{fig:cluster-palm-a}
	\end{subfigure}
	\hspace*{\fill} 
	\begin{subfigure}{0.22\textwidth}
	\begin{center}
		\includegraphics[width=0.9\linewidth]{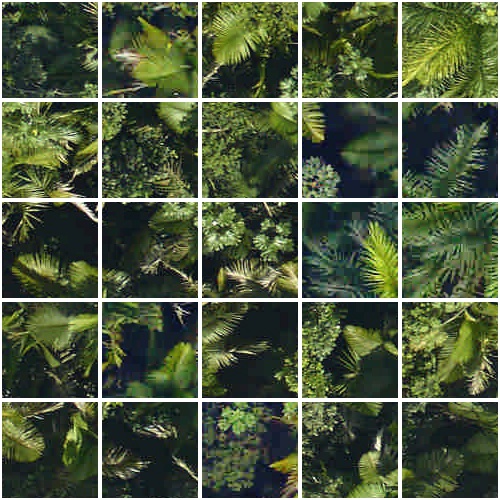}
	\end{center}
		\caption{Cluster 8} \label{fig:cluster-palm-b}
	\end{subfigure}
	\hspace*{\fill} 
	\begin{subfigure}{0.22\textwidth}
	\begin{center}
		\includegraphics[width=0.9\linewidth]{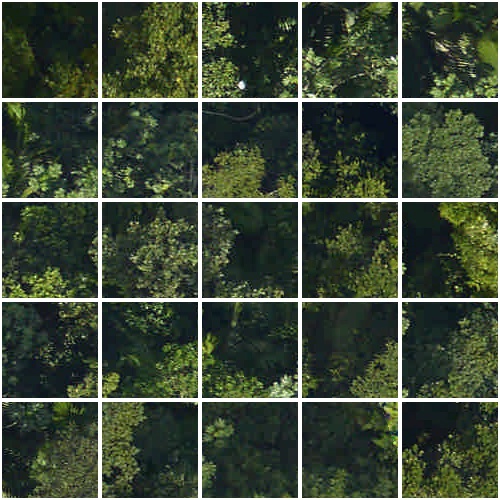}
	\end{center}
		\caption{Cluster 6} \label{fig:cluster-palm-c}
	\end{subfigure}
	\hspace*{\fill} 
	\begin{subfigure}{0.22\textwidth}
	\begin{center}
		\includegraphics[width=0.9\linewidth]{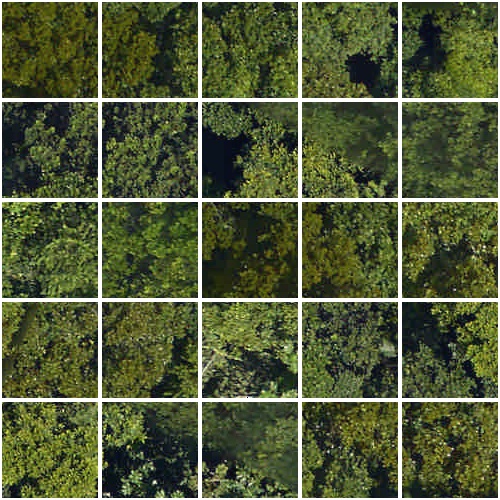}
	\end{center}
		\caption{Cluster 18} \label{fig:cluster-palm-d}
	\end{subfigure}
	\caption{Cluster visualization} 
	\label{fig:cluster-visualization}
\end{figure}

Based on the computed {\em palm relevance}, these twenty clusters were divided into two classes: {\em palm} or {\em non-palm}. In Figure~\ref{fig:cluster-density}, it is shown that two clusters stand out with very high palm relevance, as compared with the rest of the clusters. Therefore, we simply labeled these two clusters (ID: 1, 8) with the highest palm relevance as ``palm clusters'', and labeled the others as ``non-palm clusters''. Figure \ref{fig:cluster-visualization} displays randomly selected image patches from palm clusters and non-palm clusters. It can be seen that images from palm clusters (clusters 1 and 8) all ``caught a glimpse'' of a palm canopy, while patches from non-palm clusters (e.g. clusters 6 and 18) mostly showed other tree species. 

\textbf{Simplified pixel-wise classification.} With the labeled prototypes as labeled training data, we built a pixel-wise classifier using a conventional predictive modeling framework. More specifically, we trained an XGBoost classifier \cite{chen2016xgboost} over HOG features (Histogram of Oriented Gradients) \cite{dalal2005histograms}. The choice of using HOG features for the construction of this simplified classifier was due to the goal of achieving a high-resolution, pixel-wise predicted palm distribution. The VGG-19 features used in Step 2 of the APL framework have excellent discriminative power. However, the computational burden of processing VGG-19 features for sliding windows over a large volume of images is very heavy compared with other shallow features. Calculating HOG features, for example, is much faster ($\sim$100 times) than deriving VGG-19 features. For the classifier, we adopted XGBoost because it is capable to learning nonlinear functions yet can be scaled up to accommodate a large volume of data. In a sliding window of fixed size (step size = $10$ \texttt{px} $\times$ $10$ \texttt{px}, window size = $100$ \texttt{px} $\times$ $100$ \texttt{px}), dense HOG features were extracted from highly overlapped local patches of each test image. We then applied the learned XGBoost classifier to support efficient prediction on each patch. Finally, we averaged the predictions on overlapped areas into a consolidated distribution prediction. 

\begin{figure}[h]
    \centering
	\begin{subfigure}{0.23\textwidth}
	\begin{center}
		\includegraphics[width=0.9\linewidth]{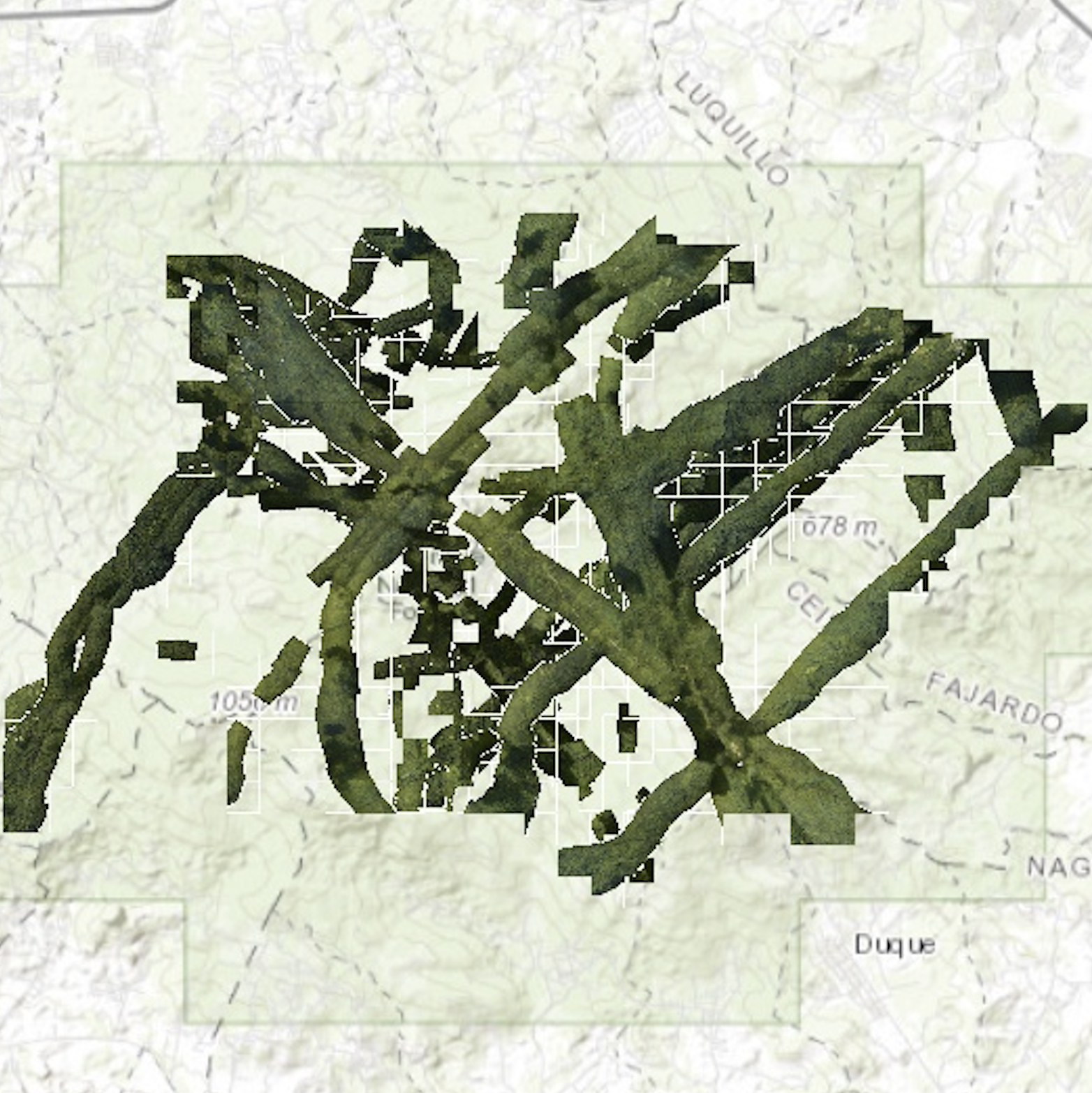}
	\end{center}
		\caption{Images over forest} \label{fig:dens-a}
	\end{subfigure}
	\hspace*{\fill} 
	\begin{subfigure}{0.23\textwidth}
	\begin{center}
		\includegraphics[width=0.9\linewidth]{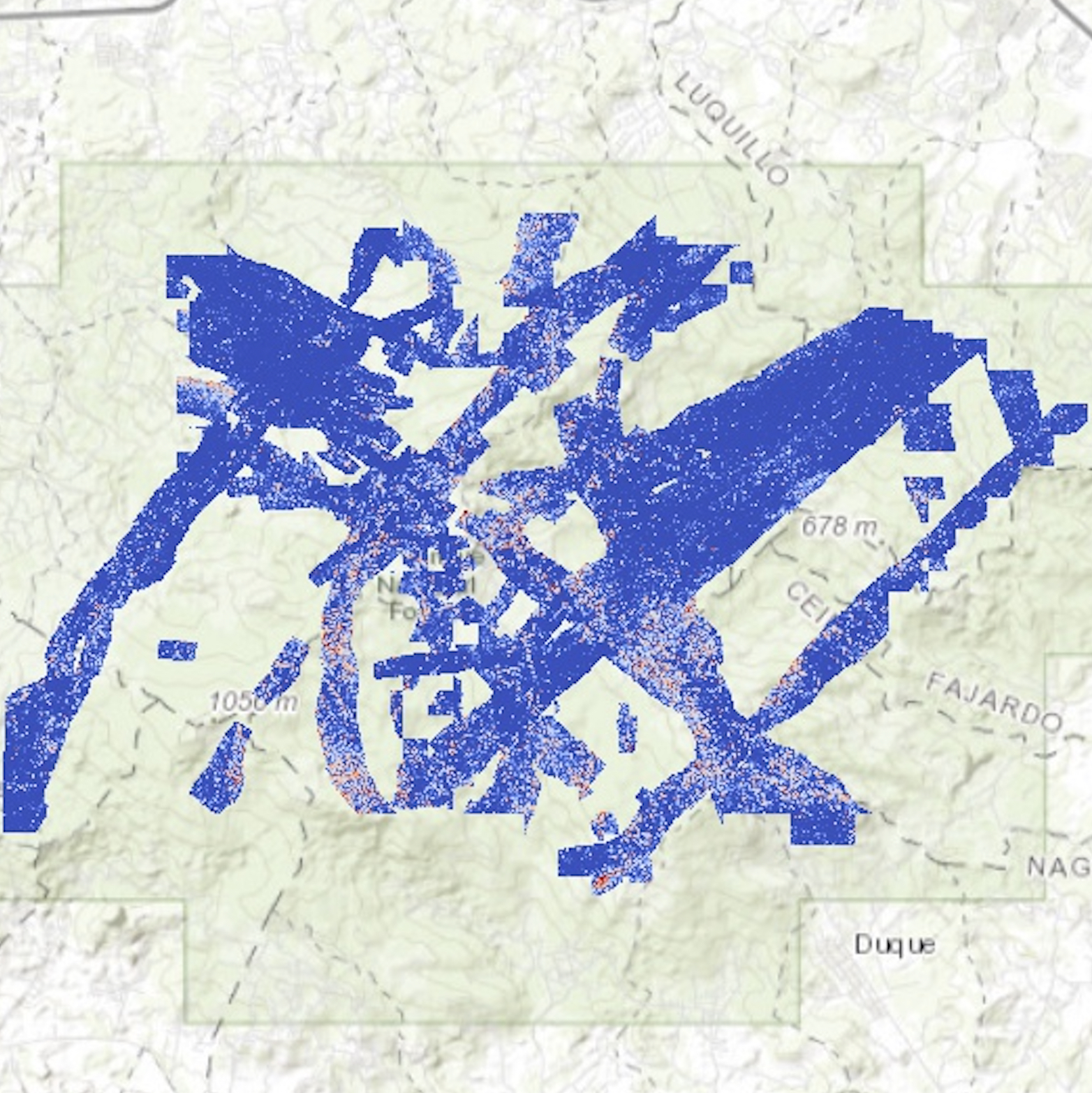}
	\end{center}
		\caption{Predictions over forest} \label{fig:dens-b}
	\end{subfigure}
	\hspace*{\fill} 
	\begin{subfigure}{0.23\textwidth}
	\begin{center}
		\includegraphics[width=0.9\linewidth]{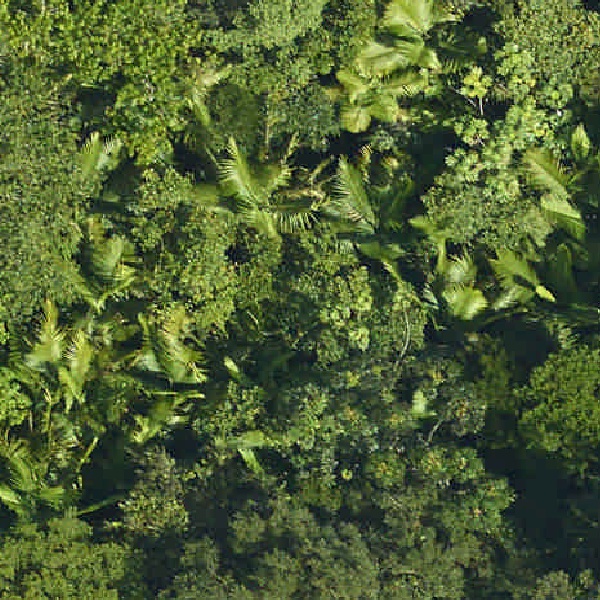}
	\end{center}
		\caption{Zoomed-in image} \label{fig:dens-c}
	\end{subfigure}
	\hspace*{\fill} 
	\begin{subfigure}{0.23\textwidth}
	\begin{center}
		\includegraphics[width=0.9\linewidth]{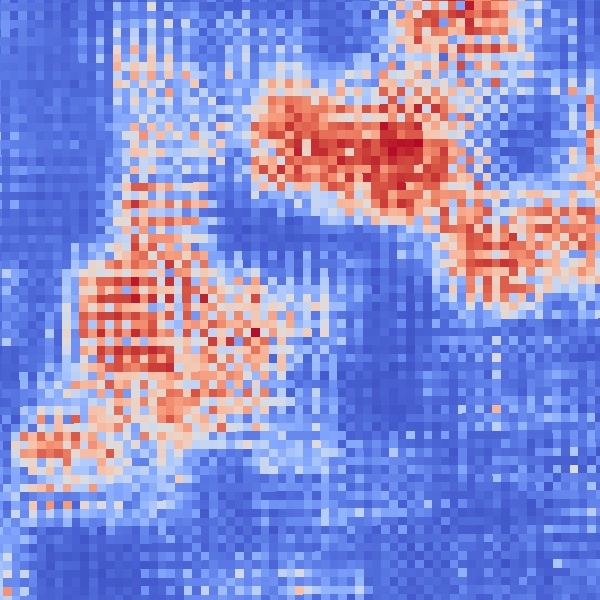}
	\end{center}
		\caption{Zoomed-in prediction} \label{fig:dens-d}
	\end{subfigure}
	\caption{Palm distribution estimation results} 
	\label{fig:prediction-results}
\end{figure}

{\bf Palm distribution estimation of the El Yunque rainforests.} Using the developed APL workflow, we estimated the palm distribution over the El Yunque rainforests using the aerial images captured by NASA Goddard's LiDAR, Hyperspectral, and Thermal (G-LiHT) Airborne Imager. To do this, we used the four aerial images with LFDP labels and another six unlabeled images as the training set and estimated palm distribution by applying the derived final simplified classifier to the rest of the images. Figure \ref{fig:prediction-results} displays our palm distribution prediction over all the aerial images. Figure \ref{fig:dens-b} shows the distribution prediction at the landscape scale (Figure \ref{fig:dens-a}), and Figure \ref{fig:dens-d} displays the prediction details on one zoomed-in image (Figure \ref{fig:dens-c}).

{\bf Performance evaluation setup.} To evaluate the performance of the proposed APL framework, we validate the workflow for calculating palm distribution on a small subset using two different measures. The target area are the four neighbouring sample images (total size = $20,000$ \texttt{px} $\times$ $20,000$ \texttt{px}) where LFDP ground labels are provided. To start, we tessellated the target area into $20 \times 20$ square subareas of equal size $1,000$ \texttt{px} $\times$ $1,000$ \texttt{px}, and randomly divided them into training (75\%) and an independent test (25\%) set. Following the proposed workflow, a predictor was trained on the training set which provides fine-grain palm canopy segmentation ($10$ \texttt{px} $\times$ $10$ \texttt{px}). 

\begin{figure}
	\begin{subfigure}{0.33\textwidth}
		\includegraphics[width=1\linewidth]{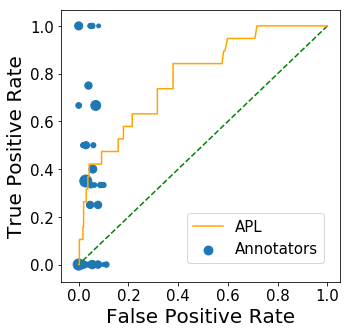}
		\caption{ROC} \label{fig:eval-1a}
	\end{subfigure}
	\hspace*{\fill} 
	\begin{subfigure}{0.67\textwidth}
		\includegraphics[width=1\linewidth]{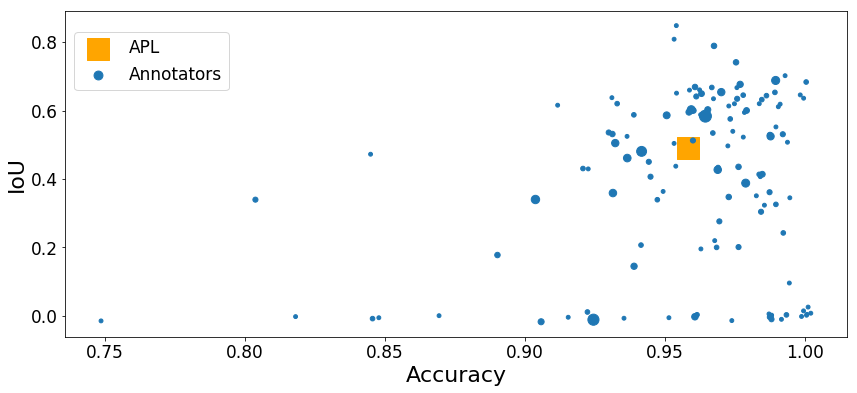}
		\caption{Accuracy vs. IoU} \label{fig:eval-1b}
	\end{subfigure}
	\caption{Performance evaluation. Left: ROC curve based on ground labels. Right: Plot for IoU vs. Accuracy for MTurk annotators and the APL framework. Point size represents the number of assignments submitted by each annotator.} 
	\label{fig:evaluation}
\end{figure}

\textbf{Evaluation using ground-based labels.} The first performance measure was based on the ground labels, in the form of a scattered point pattern, with each point indicating a corresponding trunk location. However, due to misalignment between stems and canopies, they cannot serve as a reliable reference without preprocessing. In the experiment, the scattered locations of ground labels were mapped onto a raster (matrix of cells) cell size $10$ \texttt{px} $\times$ $10$ \texttt{px}. In each cell with observed trunk locations, a majority vote of the ground labels was taken to decide the label for the cell. This allowed us to compare the resulted raster with binary labels with the map of estimated palm distribution. As the raw palm distribution predictions can take any value between 0 and 1, a continuous ROC curve was calculated by smoothly changing the discrimination threshold. For human annotators, the false positive rate and true positive rate using ground labels were also calculated in the same way, and plotted in Figure \ref{fig:eval-1a}.

\textbf{Evaluation using human annotated images.} Considering the misalignment between tree locations and canopies, we also built a segmentation task on Amazon MTurk to establish human reference data for a small set of images. With each subarea image ($1,000$ \texttt{px} $\times$ $1,000$ \texttt{px}) as an assignment, online MTurk workers segmented palm tree canopies by drawing closed polygons. For better reference quality, they received an informative tutorial about distinguishing palm trees in rainforests before starting the task. Moreover, in the submission collection process, we deleted apparently invalid segmentation submissions and republished the corresponding assignment until five valid submissions were collected for each assignment.

\begin{figure}
    \centering
	\begin{subfigure}{0.22\textwidth}
	\begin{center}
		\includegraphics[width=0.9\linewidth]{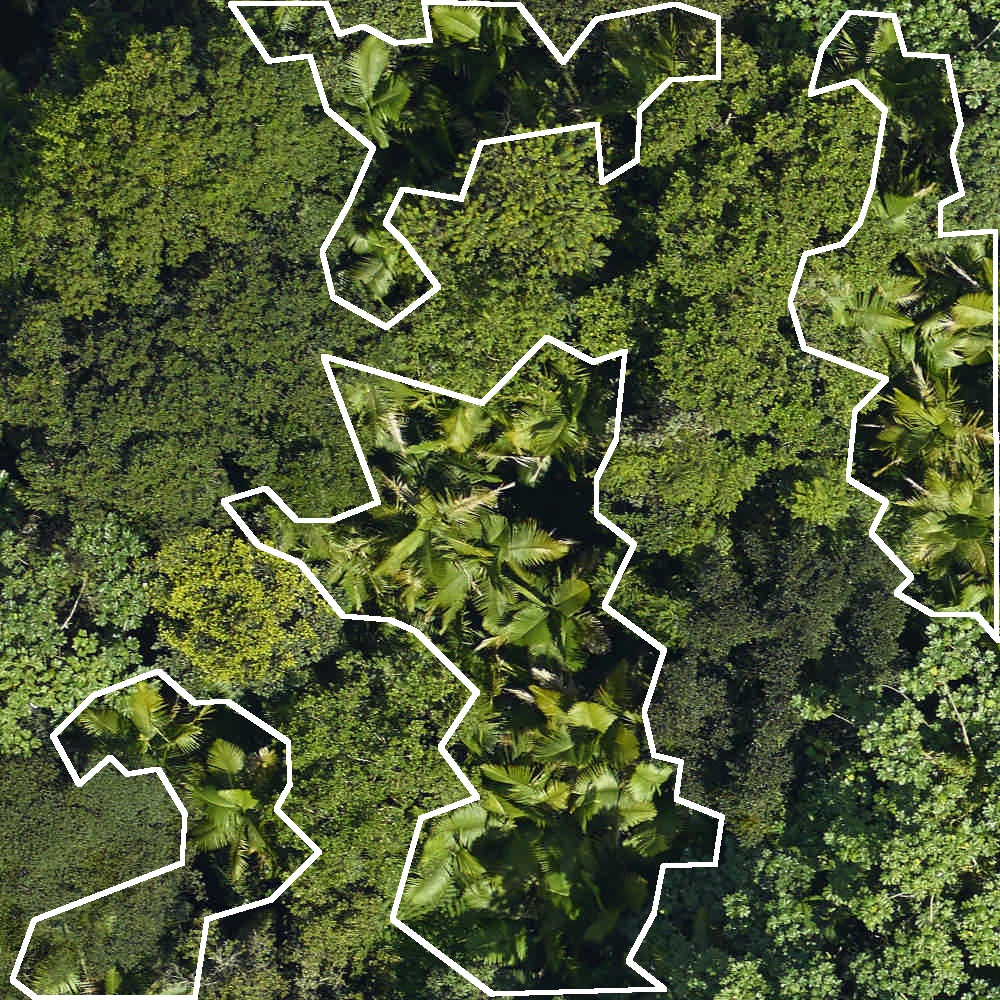}
	\end{center}
		\caption{Submission example} \label{fig:seg-1a}
	\end{subfigure}
	\hspace*{\fill} 
	\begin{subfigure}{0.22\textwidth}
	\begin{center}
		\includegraphics[width=0.9\linewidth]{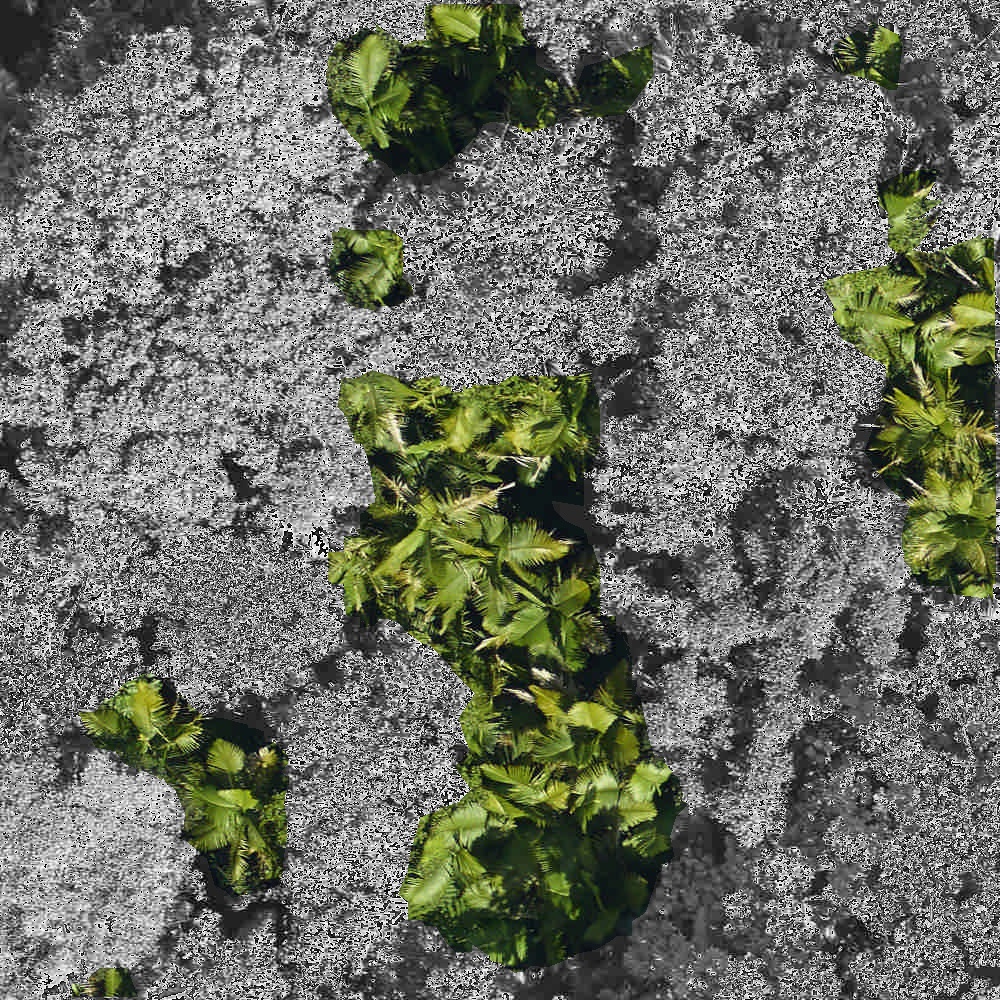}
	\end{center}
		\caption{Human reference} \label{fig:seg-1b}
	\end{subfigure}
	\hspace*{\fill} 
	\begin{subfigure}{0.22\textwidth}
	\begin{center}
		\includegraphics[width=0.9\linewidth]{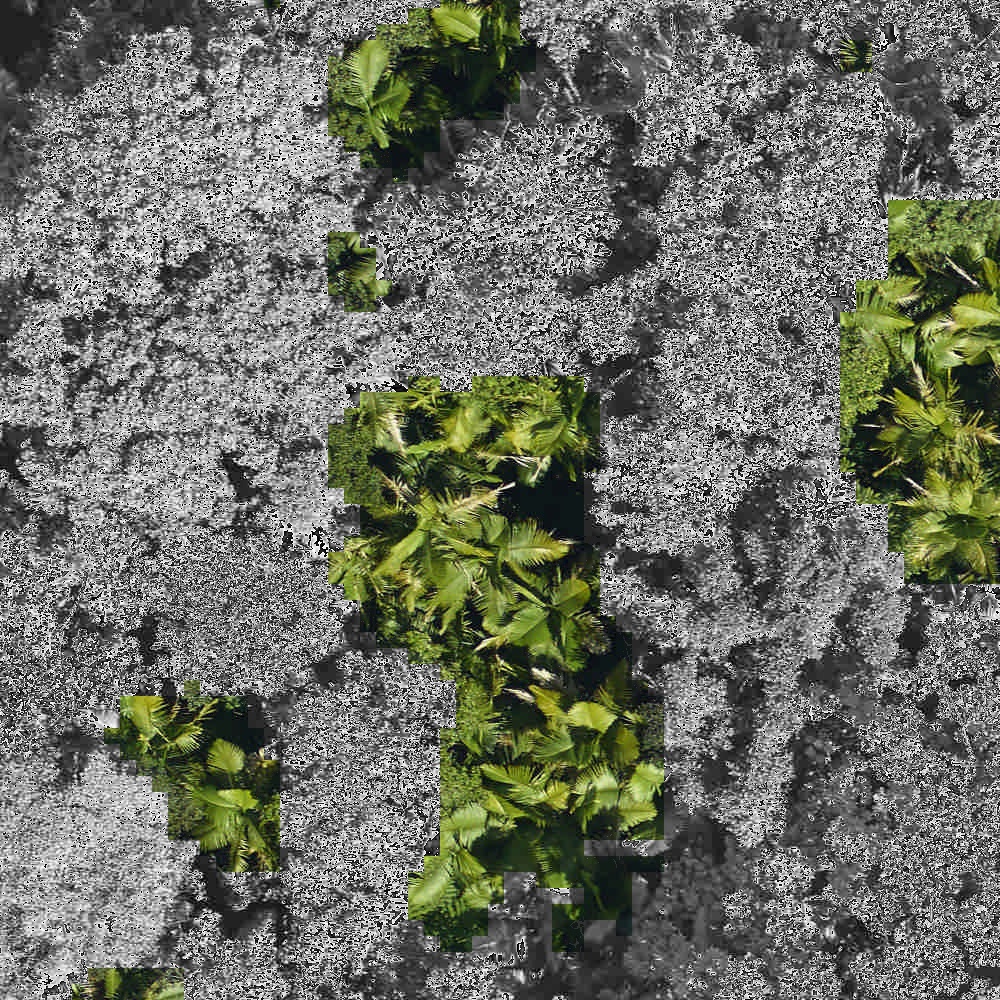}
	\end{center}
		\caption{APL prediction} \label{fig:seg-1c}
	\end{subfigure}
	\hspace*{\fill} 
	\begin{subfigure}{0.22\textwidth}
	\begin{center}
		\includegraphics[width=0.9\linewidth]{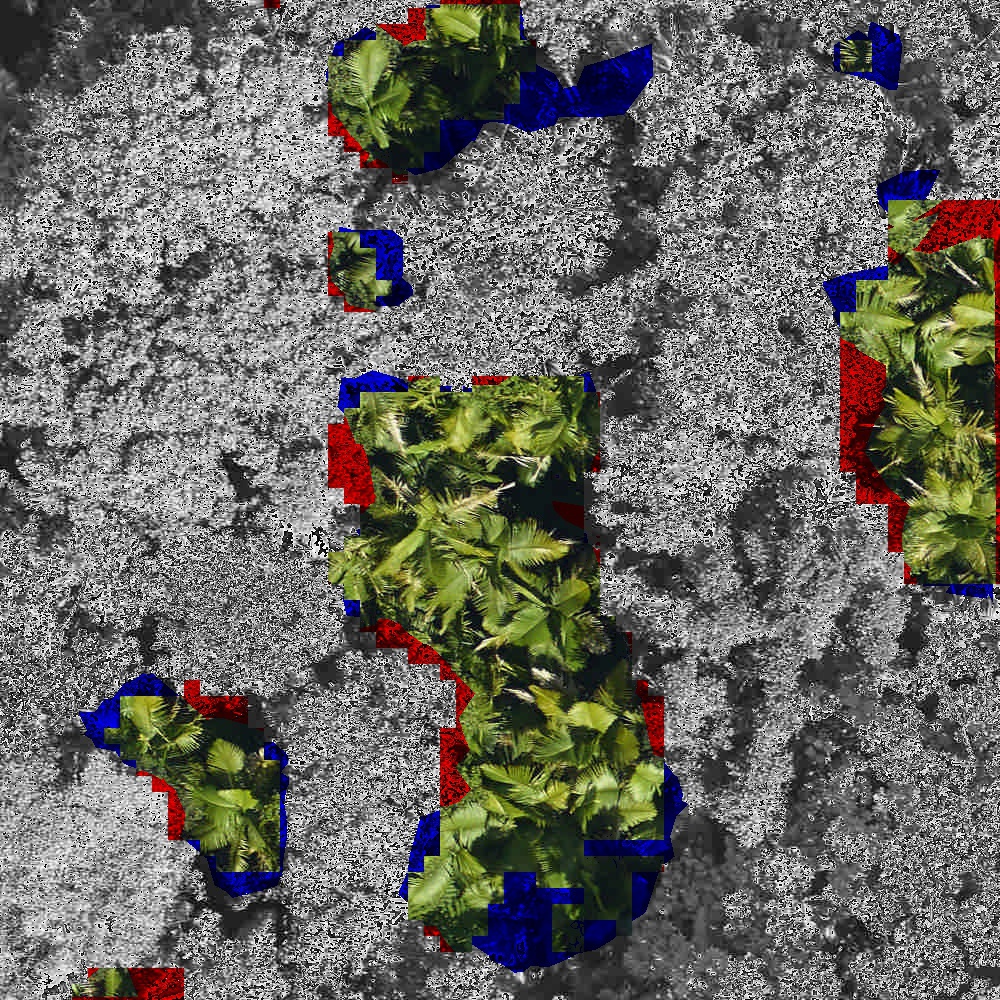}
	\end{center}
		\caption{Prediction difference} \label{fig:seg-1d}
	\end{subfigure}
	\caption{Comparing APL learning results with human annotators. (a): MTurk submission example, (b): Majority vote from five human annotators, (c): Palm prediction by the APL workflow, (d): Overlaps and differences between palm prediction by APL and human reference, where green = overlap, blue = human only, red = APL only.} 
	\label{fig:submission-example}
\end{figure}

With five human segmentation submissions on each image, we took their pixel-wise majority vote to construct human segmentation reference. Figure \ref{fig:submission-example} display a submission example, a human reference example consolidated based on five submissions, as well as the corresponding palm distribution prediction by APL, and the overlaps and differences between APL results and MTurk-generated human reference. From Figure \ref{fig:seg-1d}, differences between APL results and the human reference are mainly at the canopy boundaries, which may be due to the lack of boundary information in training set. Treating human reference as the ground truth, APL's palm prediction had a pixel-wise error rate of 95.8\% and IoU (Intersection over Union) of 0.49. Figure \ref{fig:eval-1b} compares APL with all the MTurk workers in terms of accuracy and IoU. In Figure \ref{fig:eval-1b}, the prediction performance of APL is shown to be close to the average of human annotators. While the average time an MTurk worker took to segment a $1,000$ \texttt{px} $\times$ $1,000$ \texttt{px} image was 76 seconds, it took merely 0.6 second for the simplified classifier constructed by the APL workflow to complete the same task (using a 2.4 GHz Intel Xeon processor). 

\section{Conclusion}
In this paper, we tackle the problem of weakly supervised image categorization with artificial perceptual learning (APL). Within this framework, state-of-the-art machine learning algorithms are employed as building blocks to mimic the early stage of human concept development. To validate the proposed framework, experiments were conducted over a wide-field unlabeled ecological remote sensing data, and MTurk human annotations were collected for performance evaluation. Results show our framework is able to attain human-level cognitive economy with a much lower expense.

Further work is needed to generalize the framework to multi-label classifications for other tree species. Specifically, the assumption of spatial continuity could be incorporated into the application for ecological studies. Framework extensions to other domains such as audio and natural language processing would also be of interest.

\bibliographystyle{unsrtnat}
\bibliography{refs}

\end{document}